\documentclass[letterpaper, 10 pt, conference]{ieeeconf}  % Comment this line out if you need a4paper

\IEEEoverridecommandlockouts                              % This command is only needed if 
\usepackage{amsmath} % assumes amsmath package installed
\usepackage{amssymb}  % assumes amsmath package installed
\usepackage{graphicx}
\usepackage{bbding}
\usepackage{hyperref}
\usepackage{pdfpages}
\usepackage{booktabs}

\newcommand{\myparagraph}[1]{\noindent\textbf{#1}~}
\newcounter{RNum}

\newcommand{\fref}[1]{Fig.~\ref{#1}}

\hypersetup{
    colorlinks=true     % Color of external links
}
\usepackage[font={small}]{caption}

\graphicspath{{Images/}}

\title{\LARGE \bf
AirShot: Efficient Few-Shot Detection for Autonomous Exploration 
}

\author{Zihan Wang$^{1, 2}$, Bowen Li$^{2}$,  Chen Wang$^{3}$, and Sebastian Scherer$^{2}$% <-this % stops a space
\thanks{$^{1}$Institute for Imaging, Data and Communications, School of Engineering, The University of Edinburgh, UK
        {\tt\small zwang114@ed.ac.uk}}%
\thanks{$^{2}$AirLab, Robotics Institute, Carnegie Mellon University, Pittsburgh, PA 15213, USA 
        {\tt\small zihanwa3@cs.cmu.edu; bowenli2@andrew.cmu.edu; basti@andrew.cmu.edu}}%
\thanks{$^{3}$Spatial AI \& Robotics (SAIR) Lab, Institute for Artificial Intelligence and Data Science, Department of Computer Science and Engineering, University at Buffalo, NY 14260, USA. 
        {\tt\small chenw@sairlab.org}}%
}

\begin{document}
\maketitle
\thispagestyle{empty}
\pagestyle{empty}
%%%%%%%%%%%%%%%%%%%%%%%%%%%%%%%%%%%%%%%%%%%%%%%%%%%%%%%%%%%%%%%%%%%%%%%%%%%%%%%%
\begin{abstract}
Few-shot object detection has drawn increasing attention in the field of robotic exploration, where robots are required to find unseen objects with a few online provided examples. Despite recent efforts have been made to yield online processing capabilities, slow inference speeds of low-powered robots fail to meet the demands of real-time detection-making them impractical for autonomous exploration. Existing methods still face performance and efficiency challenges, mainly due to unreliable features and exhaustive class loops. In this work, we propose a new paradigm AirShot, and discover that, by fully exploiting the valuable correlation map, AirShot can result in a more robust and faster few-shot object detection system, which is more applicable to robotics community. The core module Top Prediction Filter (TPF) can operate on multi-scale correlation maps in both the training and inference stages. During training, TPF supervises the generation of a more representative correlation map, while during inference, it reduces looping iterations by selecting top-ranked classes, thus cutting down on computational costs with better performance. Surprisingly, this dual functionality exhibits general effectiveness and efficiency on various off-the-shelf models. Exhaustive experiments on COCO2017, VOC2014, and SubT datasets demonstrate that TPF can significantly boost the efficacy and efficiency of most off-the-shelf models, achieving up to 36.4\% precision improvements along with 56.3\% faster inference speed. Code and Data are at: \url{https://github.com/ImNotPrepared/AirShot}.

% Key Word: Few-shot Detection, Robot Exploration.

\end{abstract}

% However, the requirement of an exhaustive offline fine-tuning stage in existing methods severely hinders application in autonomous exploration of low-power robots, even current method without fine-tuning stage is suffered from performance and efficiency issue. 
% The main constraints of these method are caused by unreliable correlation map and extra but useless computational cost from class-by-class detection. To taclke with these problems, we proposed a new contrastive module in training which  performs at correlation map level without any region proposals, and utilize the information in inference to pre-select class which is not likely to appear in the query image thus significantly improve the efficiency, achieved up to \textit{20\%} improvements in performance with \textit{40\%} faster sampling speed. It is a plug-and-play module with great generalization which can be applied to any existing meta-learning based few-shot object detection design. The code and pre-trained module can be found at [link]

%%%%%%%%%%%%%%%%%%%%%%%%%%%%%%%%%%%%%%%%%%%%%%%%%%%%%%%%%%%%%%%%%%%%%%%%%%%%%%%%
\section{INTRODUCTION}

Few-shot object detection (FSOD)~\cite{fan2020few,Hu2021CVPR,kang2019few,wang2019meta,wang2020frustratingly} aims to detect objects out of the base training set with only a few support examples per novel class. 
This research area has garnered increasing interest in the robotics community as it plays an essential role in autonomous exploration~\cite{chen_tro,wang2020visual,li2022airdet,kim2022robotic} where robots are expected to detect novel objects in an unknown environment while only a limited number of examples can be provided online by a human operator~\cite{kim2022robotic}. 

%The primary challenge in this context remains twofold: firstly, effectively leveraging the limited information available in a few-shot setting to provide effective supervision; and secondly, conducting efficient inference within the inherent computational constraints of low-powered robotic systems. 

However, most FSOD methods~\cite{kang2019few,wang2019meta,wang2020frustratingly,li2021few,yang2022efficient,fan2021generalized,qiao2021defrcn,sun2021fsce,wu2020multi,zhang2021accurate} cannot be directly applied to real-world robots because they are computationally heavy \cite{li2022airdet}.
One of the reasons is that they require an offline fine-tuning stage on novel classes, which is impractical for \textit{robot online} exploration.
% , since the computational and time cost of fine-tuning is unaffordable for low-powered robots~\cite{li2022airdet}. 
Even the models~\cite{fan2020few, li2022airdet} that can work without fine-tuning, still have mainly two drawbacks hindering their effectiveness in robotics. Firstly, two-stage detection frameworks~\cite{fan2020few,li2022airdet} heavily rely on region proposals to produce final predictions, which can easily fall short due to inaccurate proposals. Yet, producing high-quality region proposals for novel classes is difficult since their semantic knowledge is not learned in the feature extractor~\cite{fan2020few,zhang2020cooperating}. This issue becomes even more pronounced in the context of robot exploration, where the few-shot setting typically provides limited information. Secondly, the inference stage of previous designs~\cite{fan2020few,Hu2021CVPR,li2022airdet} follows the most exhaustive paradigm, \textit{i.e.}, running loop inference on all potential novel classes. Such a design imposes a significant computational burden, particularly on low-powered robotic platforms, making it inefficient and impractical for real-world applications.

\begin{figure}[!t]
    \centering
    \includegraphics[width=0.45\textwidth]{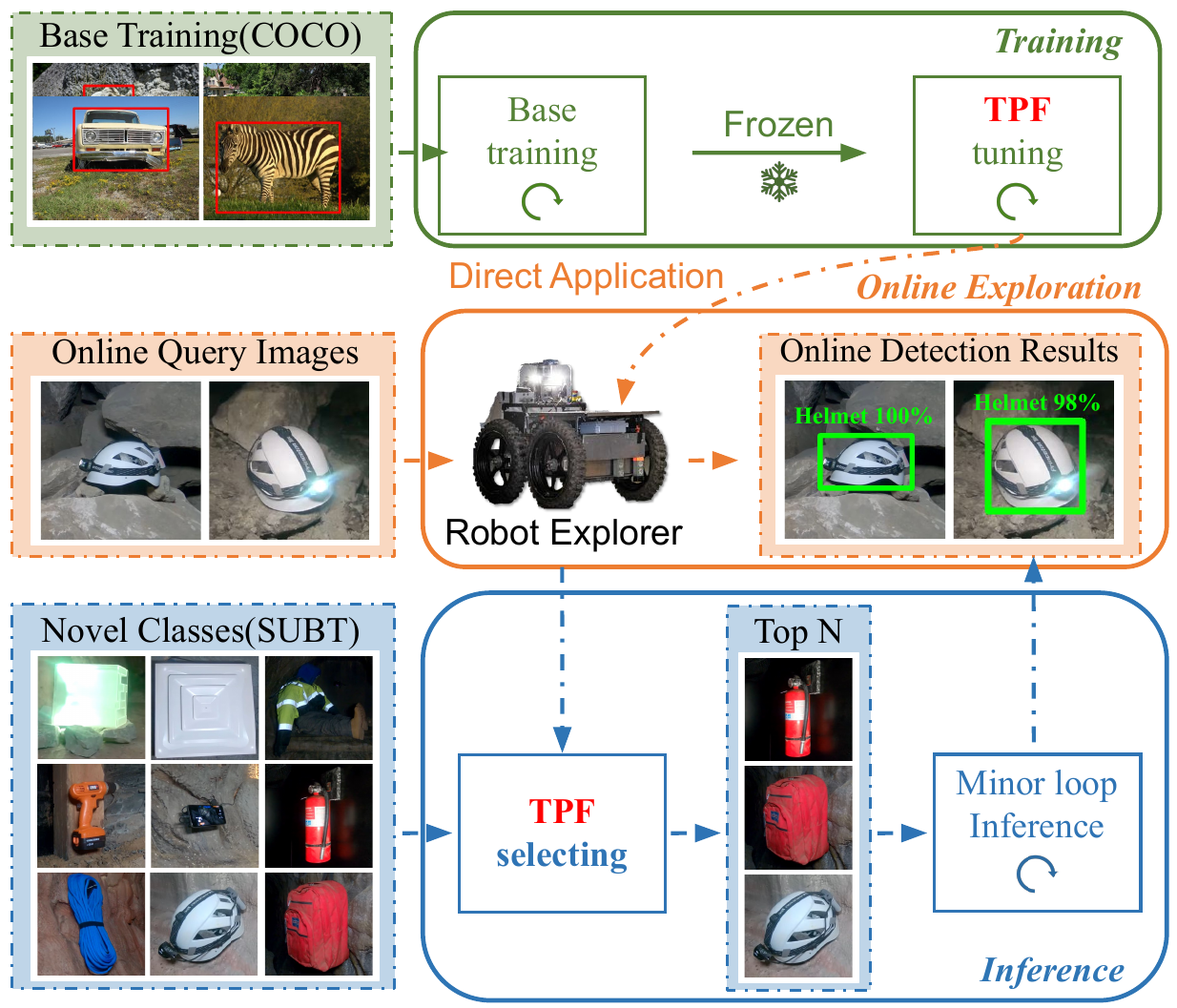}
    \caption{Application sketch of AirShot. During training, we use TPF to increase the representation capability of correlation maps. When directly applied to Robot Explorer, TPF conducts pre-selection to enable minor loop inference instead of traditional full loops.}
    \label{fig:1}
\end{figure}

%The primary challenge in this context remains twofold: firstly, effectively leveraging the limited information available in a few-shot setting to provide effective supervision; and secondly, conducting efficient inference within the inherent computational constraints of low-powered robotic systems. 
In this work, we find that a valuable correlation map can be used to solve the two problems in a unified manner. As shown in \fref{fig:1}, we introduce a simple yet effective module, Top Prediction Filter (TPF) that operates on the correlation maps during both training (green) and inference (blue) stages. % for the aforementioned two issues, respectively.

\myparagraph{Training Stage} The supervision signals of most previous work~\cite{fan2020few,kang2019few,wang2019meta,wang2020frustratingly} are provided for generating, classifying, and regressing the region proposals, which are primarily for \textit{local} anchors.
However, the semantic cues directly from \textit{global} correlation maps are overlooked.
Thus our insight is intermediate supervision on the \textit{global} feature could be beneficial. Specifically, TPF is trained to infer the existence of the category of support image directly from the global correlation maps via contrastive loss~\cite{hadsell2006dimensionality}, which generates a higher quality and more reliable correlation map.

\myparagraph{Inference Stage} Most existing methods~\cite{fan2020few, wang2019meta, wang2020frustratingly, li2022airdet} perform a full loop over all potential novel classes, which is not only computationally intensive—making them impractical for tasks with low-powered robots—but also leads to slower inference speeds, failing to meet the demands of real-time detection.
We observed that not all the potential novel classes will appear on the test images, thus the exhaustive loops have wasted much time on the absent classes.
This inspired us to design a lightweight pre-selector to avoid the extra cost during inference. In AirShot, TPF can conduct a rough classification, directly inferring the existence and offering confidence scores of certain categories.
For those categories with high confidence scores (top prediction classes), they will be fed to the following networks for finer classification and precise location. In contrast, those with low scores are considered unlikely to appear thus discarded.
% w/o causing performance descent.

%Surprisingly, we found that such an efficient pre-selection process can also result in more accurate classification since the discarded nonexistent classes will not cause misclassification of the detection heads.

Our TPF can actively provide supervision on correlation maps for more reliable and representative features during training, which effectively extract additional valuable information from limited data in few-shot scenarios for robot exploration. Moreover, its pre-selection ability significantly reduces the computational cost of low-powered robot system and further enables time-efficient detection. This dual functionality allows our approach to substantially improve the efficiency and effectiveness of the off-the-shelf models. 
In general, we summarize our contribution as twofold:

\begin{itemize}
    \item We proposed a new model AirShot that fully exploits the valuable knowledge in the correlation map. Its core module TPF, a plug-and-play design, also works generally for various FSOD models~\cite{fan2020few,li2022airdet}. The efficiency and effectiveness brought by TPF offer substantial advancements in robot autonomous exploration task. 

    \item We comprehensively test AirShot across two widely used datasets, MS-COCO and Pascal VOC. AirShot yields a 36.4\% performance improvement and a 56.28\% reduction in computational costs, which demonstrate a new SOTA in the field of FSOD that requires no fine-tuning. To demonstrate the effectiveness for real-world environments, we also tested our system with a challenging FSOD dataset, collected from the DARPA Subterranean (SubT) Challenge by our team.

    %We collect and open-source a real, datasets with DARPA Subterranean (SubT) challenge.    The effectiveness has been proven by exhaustive experiments on MS-COCO and Pascal VOC datasets.

\end{itemize}

\section{Related Work}
\subsection{General Object Detection}
Object detection~\cite{girshick2015fast, girshick2014rich, liu2016ssd, redmon2017yolo9000, ren2015faster} constitutes a pivotal challenge in the computer vision community, which aims to predict the categories and locations of predefined objects in an image. 
Modern object detection methods are primarily categorized into two classes: anchor-free and anchor-based detectors. Anchor-free detectors utilize a one-stage model structure, avoiding the explicit generation of proposal boxes. 
These methods either tackle object detection as an end-to-end regression problem, such as YOLO series~\cite{redmon2017yolo9000, redmon2016you, redmon2018yolov3} or utilize pre-defined bounding boxes to address varying object scales, as in the SSD series~\cite{liu2016ssd}. 
Anchor-based detectors~\cite{girshick2014rich}, first generate class-agnostic region proposals using Region Proposal Network (RPN). These proposals are further refined and classified into different categories through detection heads. By filtering out negative locations using RPN, anchor-based methods often achieve decent results in general detection tasks~\cite{ren2015faster, lin2017fpn}.
Both anchor-based and anchor-free methods require intensive supervision and have a fixed number of object classes post-training, making them unsuitable for tasks such as autonomous exploration, where unseen and novel objects appear dynamically. 

\subsection{Few-Shot Object Detection}
Few-shot object detection leverages the knowledge from abundant base class data to generalize to novel categories with only a few labeled examples as support.  
Two main branches are thriving in recent FSOD research, namely meta-learning-based approaches~\cite{fan2020few, wu2020multi, yan2019meta, han2021query} and transfer-learning-based approaches~\cite{wang2020frustratingly,qiao2021defrcn,sun2021fsce,zhu2021semantic,wu2021universal}.

Transfer-learning~\cite{wang2020frustratingly,qiao2021defrcn,sun2021fsce,zhu2021semantic,wu2021universal} aims to identify the best fine-tuning strategy to adapt general object detectors to novel images with limited examples. 
For instance, Wang \textit{et al.}~\cite{wang2020frustratingly}  proposed fine-tuning only the last layer. 
Wu \textit{et al.} addressed scale scarcity through manually defined positive refinement branches, such as MPSR~\cite{wu2020multi}.
Other works also explored semantic relationships between novel and base classes applying contrastive proposal encoding~\cite{sun2021fsce}.

Meta-learning-based methods aim to train meta-models on individual tasks in an episodic manner, with separate branches for extracting support information and detecting objects within query images. 
Notable contributions in this area include Meta R-CNN~\cite{yan2019meta} and Meta-DETR~\cite{zhang2022meta}.
Meta R-CNN~\cite{yan2019meta} focused on support-guided query channel attention, A-RPN~\cite{fan2020few} with novel attention RPN and a multi-relation classifier.
Meta-DETR~\cite{zhang2022meta}, the current state-of-the-art, introduced a correlation aggregation module to simultaneously aggregate multiple support categories to capture their inter-class correlation.
Recent developments include support-query mutual guidance~\cite{zhang2021accurate}, context information aggregation~\cite{Hu2021CVPR}, and the construction of heterogeneous graph convolutional networks on proposals~\cite{han2021query}. Transformer-based method FCT~\cite{han2022few} proposed a model with three interaction stages between query and support in the backbone and one additional interaction stage in the multi-relation detection head, which computes similarities between support and query features to output the final detection results.

Despite recent progress, few-shot learning still faces several challenges. Region-based detection
frameworks rely on region proposals to produce final predictions, however, it is not easy to produce high-quality region proposals for novel classes with limited supervision under the few-shot detection setups~\cite{zhang2022meta} and the need for re-training with new categories also limits the development. 
Therefore, a more robust, flexible, and efficient few-shot object detection method that can better adapt to different scenarios and novel categories without fine-tuning stage is needed.

\subsection{Multi-Scale Feature Extraction}
The utilization of multi-scale features in the detection of objects with varying scales has been extensively explored~\cite{liu2016ssd, redmon2017yolo9000, Kong2016HyperNet, li2017fssd, lin2017fpn, Shen2017dsod}, which has been proved necessary for small object detection. 
For example, the feature pyramid network (FPN) employed a multi-scale feature map for detection. 
To exploit information in multi-scale features, FSSD~\cite{li2017fssd} and AirDet~\cite{li2022airdet} proposed a multi-scale feature fusion module. While others~\cite{sun2021fsce, zhang2021accurate,zhu2021semantic, lin2017fpn} employed all scales from FPN and implemented detection on each scale parallel. Granted, a multi-scale feature is necessary to effectively detect small novel classes, but existing methods are cumbersome when running it for all classes. In contrast, our model AirShot can utilize multi-scale features from correlation map more efficiently.

\subsection{Few-Shot Detection without fine-tuning}
Although FSOD has made significant progress recently, most existing methods~\cite{fan2020few,li2022airdet} followed the two-stage training paradigm, which requires base training and fine-tuning stages. 
However, the fine-tuning stage cannot be applied to robot online exploration due to the following concerns: 1) dynamically changing categories, 2) limited onboard computing power of the robot, 3) non-existing evaluation dataset~\cite{li2022airdet}. 
This task is important for robot exploration and perception in unseen environments, yet, only little progress has been made so far~\cite{fan2020few,li2022airdet}. 
In this work, the core module of AirDet, TPF,  acts as a simple yet effective approach to boost both the efficiency and efficacy of these methods~\cite{fan2020few,li2022airdet}.

\section{Methodology}
\subsection{Preliminary}
\subsubsection{Exploration Task}
The exploration task is executed under a few-shot object detection setting. 
Due to the unseen nature of objects during exploration, it requires model trained on base classes $\mathcal{C}_{\mathrm{b}}$ can detect novel classes in $\mathcal{C}_{\mathrm{n}}$, satisfying $\mathcal{C}_{\mathrm{n}} \cap \mathcal{C}_{\mathrm{b}} = \emptyset$. 
To save human effort, only a few annotated images ($k$-shot samples per novel class) are available during online exploration. 
Note that fine-tuning is not practicable under this setting.
During the exploration, the robot will continuously collect images. 
The human user needs to annotate the novel objects first and provide them back to the robot explorer. 
Then the robot explorer should detect unseen objects by observing the surrounding environment. The main challenge of autonomous exploration task exhibits in two main perspectives: one is how to fully utilize limited information in few-shot setting to provide effective supervision, the other is how to run efficient inference when the inherit computation cost is limited for low-powered robots.

\subsubsection{AirDet Review}
\begin{figure}[!t]
    \centering
    \includegraphics[width=0.45\textwidth]{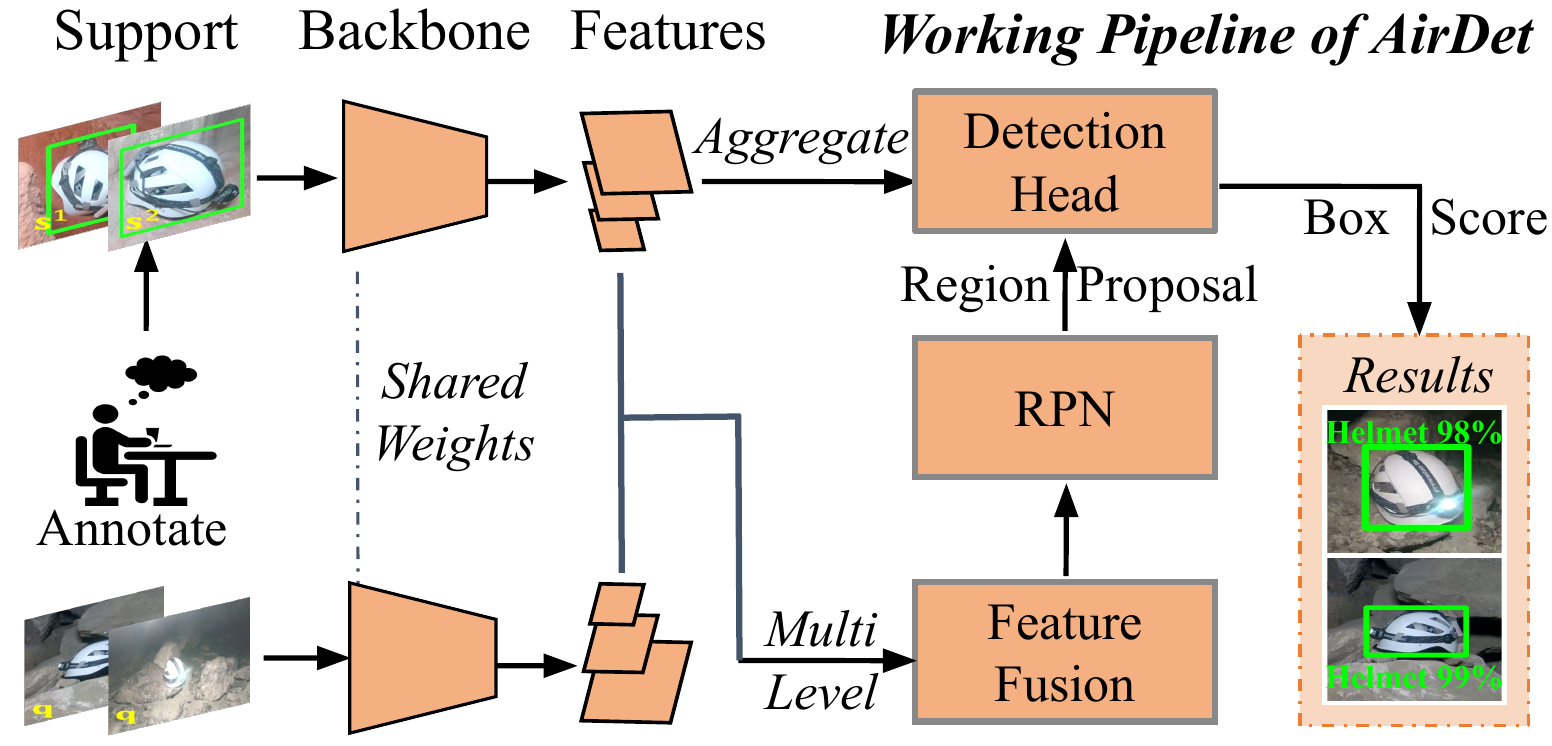}
    \caption{Working pipeline of AirDet. AirDet includes 3 modules, i.e., the shared backbone, feature fusion module for region proposal and shots aggregation, plus relation-based detection head.}
    \label{AirDet}
\end{figure}

AirDet~\cite{li2022airdet} is a meta-learning-based few-shot object detector as shown in Fig. \ref{AirDet}. 
It shows favorable performance without finetuning thanks to its class-agnostic design. 
With the proposed spatial relation and channel relation, AirDet constructs a support-guided cross-scale (SCS) module as a feature fusion of region proposals, a global-local relation network for shots aggregation, and a prototype relation embedding for precise localization. 
During exploration, a few prior raw images containing novel objects are sent to a human user for annotation as support images. % ($\mathbf{s^1}$ and $\mathbf{s^2}$). 

The support images and the new query image ($\mathbf{q}$) perceived by the explorer are fed into a shared backbone. 
Then SCS accepts extracted multi-level support and query features from different backbone blocks (ResNet~\cite{he2016deep} 2, 3, and 4 blocks) as input. 
The SCS module utilizes the relation between query and support multi-scale features to generate fused correlation maps for region proposals. 
Finally, the region proposal and aggregated class prototype from the shots aggregation module are fed to the relation-based detection head for fine classification and regression.

\begin{figure*}[t]
    \centering
    \includegraphics[width=\textwidth]{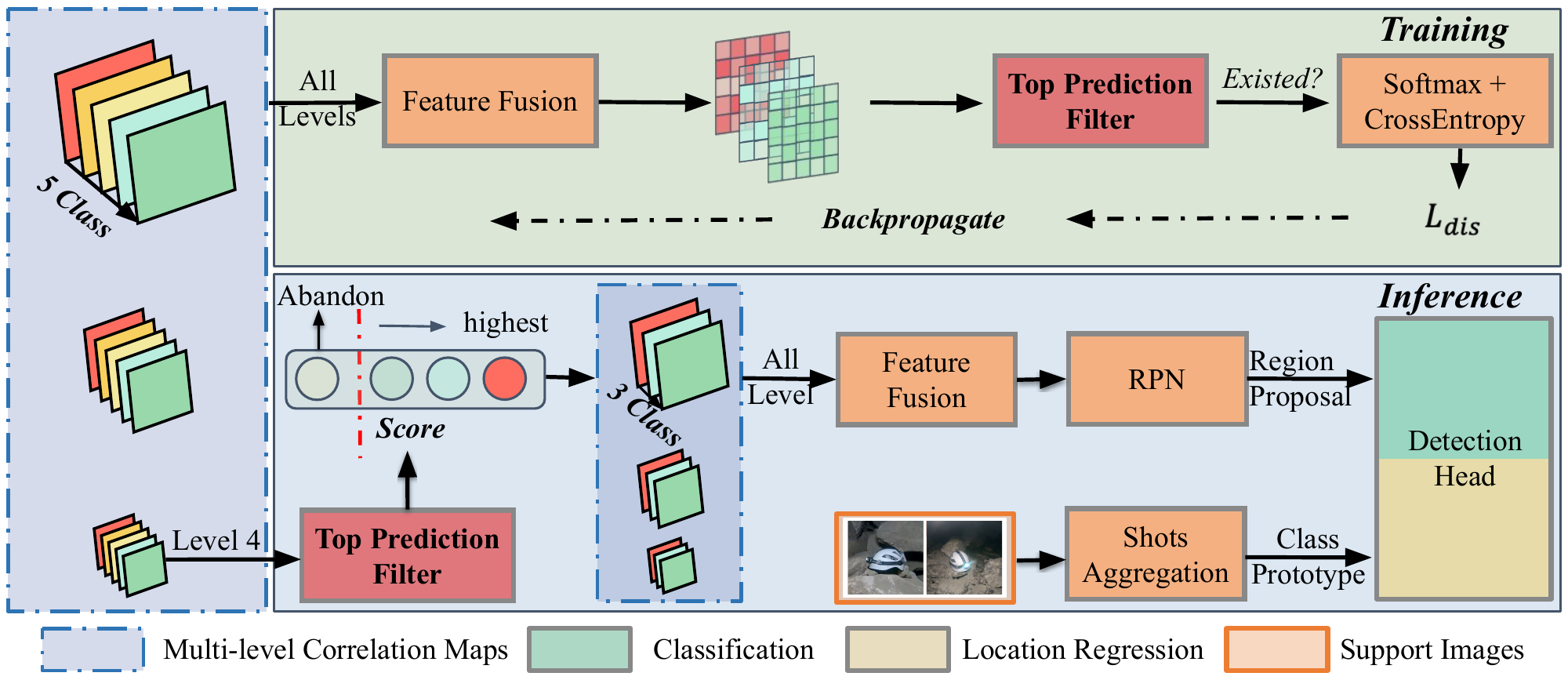}
    \caption{Detailed working illustration of AirShot in training and inference stage. We adopt the backbone design of AirDet which contains backbone feature extractor, SCS for feature-fusion, relation-based shots aggregation and location regression. }
    \label{Overall}
\end{figure*}

\subsection{Overview of Top Prediction Filter (TPF)}
\begin{figure}[!t]
    \centering
    \includegraphics[width=0.45\textwidth]{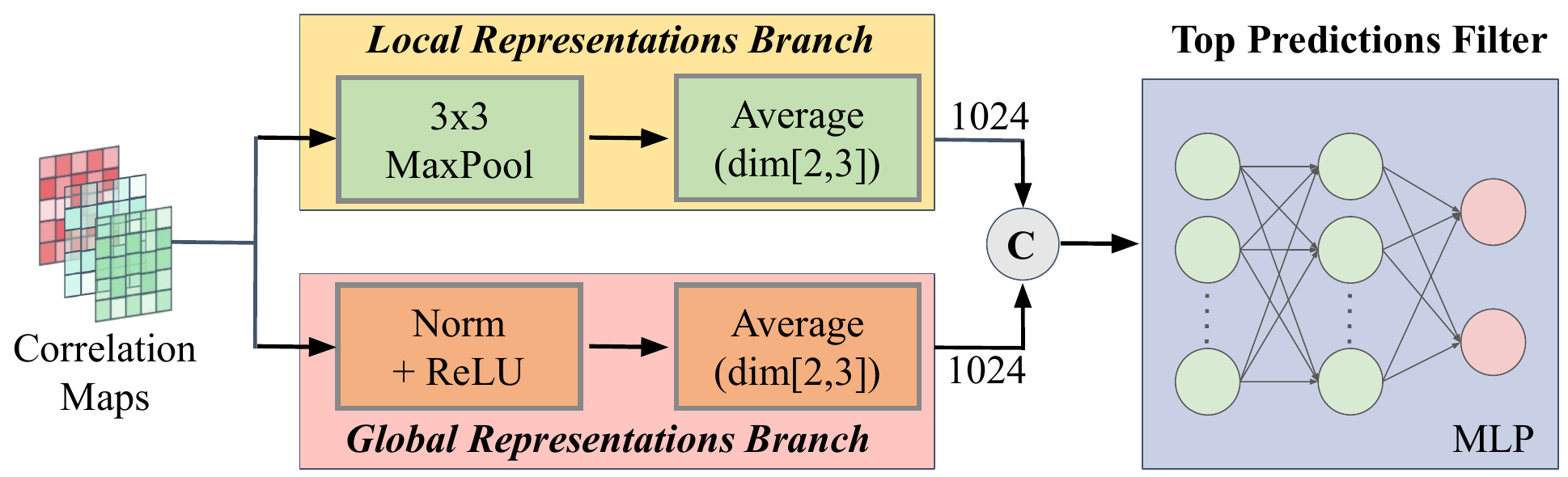}
    \caption{Network architecture of TPF. We design two branches for global representation and local representation separately. Then the concatenated representation will be fed into a 3-layer MLP.}
    \label{TPF}
\end{figure}

Despite great performance, we observe that the correlation map in AirDet is not reliable enough. 
Moreover, like various other FSOD methods~\cite{fan2020few, li2022airdet, fan2021generalized}, it has an efficiency problem, i.e., full inference loops. 
As in Table~\ref{time}, the backbone feature extraction runs fast, feature fusion (SCS in our case), RPN, and detection head occupy a majority of computational cost. 
\begin{table}[ht]
\centering
\caption{Time Consumption for Modules}
\label{time}
\begin{tabular}{c|c|c|c|c}
\toprule
  & Backbone & SCS & RPN & Detection Head   \\
\midrule
Time($s$) & 0.013  & 0.099  & 0.115 & 0.506\\
Proportion($\%$)  & 1.81 & 13.55 & 15.65 & 68.99 \\
\toprule
\end{tabular}
\end{table}
Intuitively, we do not need to exhaustively search every class with full inference loops on these modules. 
Instead, we can pre-select the classes and reduce the looping iterations to improve efficiency.
To this end, we proposed a new module named Top Prediction Filter (TPF) in Fig.~\ref{TPF}, which is proven widely effective for various few-shot detectors. 
 
As shown in the top row of Fig.~\ref{Overall}, during training, TPF is injected to provide supervision on the fused correlation map to enhance the robustness of proposal generation.
As for the inference, we feed \textit{only} the level-4 correlation map into TPF to calculate the score for each class. 
The score here represents the confidence regarding the existence of a novel class. Then a list of ranked scores is responsible for selecting a reduced number of novel classes. 
All levels of correlation maps of the remaining novel classes are then sent into the second loop including a feature fusion, proposal generation, and detection head for classification and regression.

\subsection{Top Prediction Filter (TPF)}

Our main motivation is to reduce the computational burden caused by novel classes that are unlikely to appear in the query image. 
This is particularly relevant in time-consuming modules run in each iteration. The goal of TPF is thus to establish a direct mapping from correlation maps to linearly separable scores or logits. To achieve this, it is desirable to grant the model discrimination ability, which can also guide better correlation map generation during training. 

To fulfill the dual functionality, the proposed approach must adhere to the following properties:
(1) adaptable to dynamical numbers of novel classes, therefore the module should function as a binary classifier; and
(2) limited inherent computational cost, as the module itself needs to run in a full inference loop. The ideal network should strike a good balance between inherent computational cost and performance.

Motivated by the observation that stronger activation of correlation maps corresponds to higher similarity, we propose an approach that effectively leverages the abundant information captured in correlation maps, which directly maps correlation maps to linearly separable scores. 
To extract the information, we define 2 representations, global representation, and local representation, separately defined as: 

\subsubsection{Global Representation}

\begin{equation}
\mathbf{{R}_{g, i}} = \mathrm{Avg} \left ( \mathrm{ReLU} \left ( \mathrm{Norm} \left ( \mathbf{{c_i}} \right)\right)\right),
\end{equation}
where $\mathbf{{c_i}}$ is the correlation map of the ${i_{th}}$ category.

\subsubsection{Local Representation}
Due to the localized nature of the activation strength of correlation map, it is important to extract the local representation to form confidence vector:
\begin{equation}
\mathbf{{R}_{l, i}} = \mathrm{Avg} \left ( \mathrm{MaxPool} \left ( \mathbf{{c_i}}\right)\right).
\end{equation}

\subsubsection{Confidence Vector}
To better utilize both global and local representation, we simply concatenate two representations to get the fused representation as a confidence vector. 
\begin{equation}
\mathbf{{V}_{con, i}} = \mathrm{Cat} \left( \mathbf{{R}_{l, i}}, \mathbf{{R}_{g, i}} \right).
\end{equation}
Then we pass the confidence vector into a 3-layer MLP (2048, 512, 2) to get the logits for the probability of the existence of the corresponding novel class:
\begin{equation}
\mathbf{{p_i}} = \mathrm{SoftMax} \left ( \mathrm{MLP} \left ( \mathbf{{V}_{con, i}} \right)\right).
\end{equation}
Similarly, the footnote $i$ above demonstrates the corresponding variables or representations of  ${i_{\text{th}}}$ category.

\subsection{AirShot during training}

The classification supervision of most prior work~\cite{{fan2020few,Hu2021CVPR,kang2019few,wang2019meta,wang2020frustratingly}} relies on the region proposals feature extraction, neglecting the rich information from correlation maps directly. To settle unreliable correlation maps, we assume extra supervision is beneficial. Therefore during training,  AirShot utilizes the rich information in correlation maps to make direct inferences about the category's existence. We follow a contrastive strategy to determine the existence. Specifically, we feed the fused correlation map into TPF, then receive the output of the final MLP layer, followed by a cross-entropy loss to apply the contrastive learning strategy. 

As for training strategy, we first jointly train all modules in AirShot. During training TPF can act as a discriminator to encourage better correlation map generation. We then frozen the rest of the model and fine-tuned TPF separately. This is to achieve a more robust deterministic effect. 
Empirically the supervision on the fused correlation map is better than the supervision on level 4. We assume that the supervision on fused one can act across all 3 levels. Whereas in the latter setting, supervision can only function on one level. More details regarding feature level can be found in Section~\ref{different  correlation}.

\subsection{AirShot during inference}
During inference, we first utilize multi-level query features and support features to generate correlation maps, then feed the deepest correlation maps into TPF to get the logit of the final MLP layer. It is followed by a Softmax layer to normalize the score. Since we do not have positive/negative pairs during inference, we regard the logits of the single positive neuron as the final score to rank following a certain strategy.  Once the category set of top predictions is confirmed, all 3 levels of the category will be passed to the feature fusion module (in our case we adopt SCS from AirDet) to generate fused correlation maps. Then the fused correlation maps will be fed to RPN to generate region proposals. Along with the aggregated class prototype, they will be sent to detection head to infer the classification and location. However, the feature fusion module remains for the selected class sent to the detection head for comparable performance concerns.

Noticed that here we feed the deepest correlation map instead of the fused one. %Despite that the feature fusion is necessary for reliable proposal generation especially for small object,% 
We argue that the fused correlation maps are sub-optimal in our pre-select discrimination process. Moreover, this strategy offers two advantages: (1) a performance enhancement due to improved deterministic ability; and (2) the ability to skip inefficient feature fusion or scale-wise parallel computing among all the novel classes.  

Regarding strategy, we provide two types, Top $N$ and adaptive method. ``Top $N$'' refers to simply picking the categories with the $N$ highest scores ($N$ normally ranges from 5-10). However, given the fact that the number of novel classes on the query images varies, we also introduce the adaptive method, which sets a threshold and selects the class whose score is beyond the threshold. %However, we remind the reader that for the selected novel class sent to detection head, we still remain the design of feature fusion module for better performance concern. % 
This strategy successfully eliminates the abundant computational cost of the non-existent novel classes. There is little performance drop due to (1) TPF can capture most of the seen classes; and (2) the necessary feature fusion module is preserved for top predictions. Thus AirDet can achieve efficient inference with comparable performance as adopting full loop inference.

\section{Experiments}
We adopt the design from AirDet and use it as the baseline. We mainly adopt ResNet101~\cite{he2016deep} pre-trained on ImageNet as the backbone. AirShot and the baseline~\cite{fan2020few, li2022airdet} share the same supports in all the settings. We use 4 NVIDIA A100 for both training and evaluation. We mainly present $k = 1, 2, 3, 5$-shot evaluation since there are limited support samples available during real-world applications. Due to the unseen nature of objects during exploration, we only focus on novel classes throughout the experiments.
We first show the details of implantation and our evaluation metrics, then we demonstrate the results of conducting a full loop to validate our baseline enhancement, and finally, we show performance of efficient inference strategy of AirShot applying TPF.

\subsection{Implementation Details}
We first jointly train all modules in AirShot: the feature fusion part, RPN, detection head with TPF for 80K iterations from scratch. Once the training is done, we froze other parts and fine-tuned TPF separately on the training set. Note that for different evaluation settings, we apply the same trained model directly, without any further fine-tuning stage. 

\subsection{Evaluation Metrics}
We defined the strategy going through all classes as \textit{full loop}, and the loop running in our filtered categories selected by TPF as \textit{minor loop}.
We compare the performance of in-domain(COCO) and cross-domain (VOC) regarding average precision (AP). To evaluate the performance drop between minor loop utilizing TPF and full loop, we proposed a new metric called \textbf{omission rate (OR)}, defined as: 
\begin{equation}
OR =  - \frac{{AP}_{\text{full}} - {AP}_{\text{minor}}}{{AP}_{\text{full}}} \times 100\%.
\end{equation}

It is also important to evaluate recall rate of TPF for each class. Thus, we create a class-wise recall bar and show it in Fig.~\ref{recall}, where the footnote $n$ denotes the number of shots.

\subsection{Baseline Enhancement} 
In this section, we show that utilizing TPF in joint training can achieve better performance. For a fair comparison, all the experiments in this section go through a full inference loop.

\subsubsection{In-Domain Evaluation} \label{sec:in-domain-vvaluation}
\begin{table}[b]
\centering
\caption{In-domain Performance on COCO validation dataset.}
\begin{tabular}{clccc}
\toprule
 & model & AP & $AP_{50}$ & $AP_{75}$ \\
\midrule
 & A-RPN & 3.32 & 6.28 & 3.04 \\
1 shot & A-RPN+TPF  & \textbf{4.53} & \textbf{8.05} & \textbf{4.55} \\
 & AirDet & 5.41 & 10.19 & 5.64 \\
 & AirShot (Ours) & \textbf{6.28}  &\textbf{11.35} & \textbf{6.25} \\
\midrule
 & A-RPN & 4.10 & 7.72 & 3.81 \\
2 shots & A-RPN+TPF  & \textbf{4.81} & \textbf{8.70} & \textbf{4.89} \\
 & AirDet & 5.96 & 11.22 & 5.81 \\
 & AirShot (Ours) & \textbf{7.07} & \textbf{12.99} &\textbf{6.98} \\
\midrule
 & A-RPN & 5.47 & 9.80 & 5.47 \\
3 shots & A-RPN+TPF  & \textbf{5.91} & \textbf{10.53} & \textbf{5.88} \\
 & AirDet & 6.54 &12.26 & 6.33 \\
 & AirShot (Ours) & \textbf{8.01} & \textbf{14.96} &\textbf{7.87} \\
\midrule
 & A-RPN & 5.87 & 10.47 & 5.96 \\
5 shots & A-RPN+TPF  & \textbf{6.35} & \textbf{10.93} & \textbf{6.28} \\
 & AirDet & 7.91& 14.61 & 7.75 \\
 & AirShot (Ours) & \textbf{8.83} & \textbf{15.64} & \textbf{8.76} \\
\midrule
 & A-RPN & 6.01 & 10.42 & 6.16 \\
10 shots & A-RPN+TPF  & \textbf{6.49} & \textbf{11.54} & \textbf{6.64} \\
 & AirDet & 8.59 &15.15 & 8.68 \\
 & AirShot (Ours) & \textbf{9.58} & \textbf{16.64} & \textbf{9.84} \\
\bottomrule
\end{tabular}
\label{coco}
\end{table}
We first present the in-domain evaluation on the COCO benchmark in Table~\ref{coco}. Following prior works, we split the 80 classes into 60 non-VOC base classes and 20 novel classes. During training, the available images are the base class images from COCO train2014 datasets. Then the trained models are comprehensively evaluated on 5,000 images from the COCO val2014 dataset with few-shot samples per novel class as support images.

\subsubsection{Cross-Domain Evaluation}
Cross-domain performance is crucial for robotic applications as robots are often deployed to novel environments. We adopt the same model trained on COCO as in Section \ref{sec:in-domain-vvaluation} to evaluate model generalization on the PASCAL VOC dataset as in Table~\ref{val}. Note that the model is directly applied without any fine-tuning process.

\begin{table}[h]
\centering
\caption{Cross-domain performance on VOC-2012 dataset.}
\begin{tabular}{c lccc}
\toprule
 & model & AP & $AP_{50}$ & $AP_{75}$ \\
\midrule
 & A-RPN & 10.03 & 17.97 &  10.22\\
 & A-RPN + TPF & \textbf{10.73} & \textbf{18.96} & \textbf{10.41} \\
1 shot & AirDet & 10.98 & 20.36 & 10.46 \\
 & AirShot (Ours) & \textbf{11.71} & \textbf{22.68} & \textbf{11.56} \\
\midrule
 & A-RPN &  12.98 & 22.39 & 13.88 \\
 & A-RPN + TPF & \textbf{13.29} & \textbf{23.92} & \textbf{14.15} \\
2 shots & AirDet & 13.70 & 24.84 &14.36\\
 & AirShot (Ours) & \textbf{15.23} & \textbf{27.29} & \textbf{15.11} \\
\midrule
 & A-RPN & 12.60 & 20.92 &13.02 \\
 & A-RPN + TPF & \textbf{13.87} & \textbf{22.96} & \textbf{14.09} \\
3 shots & AirDet & 15.59 & 27.27 &15.82\\
 & AirShot (Ours) & \textbf{16.91} & \textbf{29.67} & \textbf{17.56} \\
\midrule
 & A-RPN & 13.19 & 21.91 & 13.65 \\
 & A-RPN + TPF & \textbf{14.48} & \textbf{23.96} & \textbf{14.57} \\
5 shots & AirDet & 16.65 &28.67&17.20 \\
 & AirShot (Ours) & \textbf{18.32} & \textbf{31.43} & \textbf{18.41} \\
\bottomrule
\end{tabular}
\label{val}
\end{table}

\begin{figure}[b]
    \centering
    \resizebox{\linewidth}{!}{
    \includegraphics{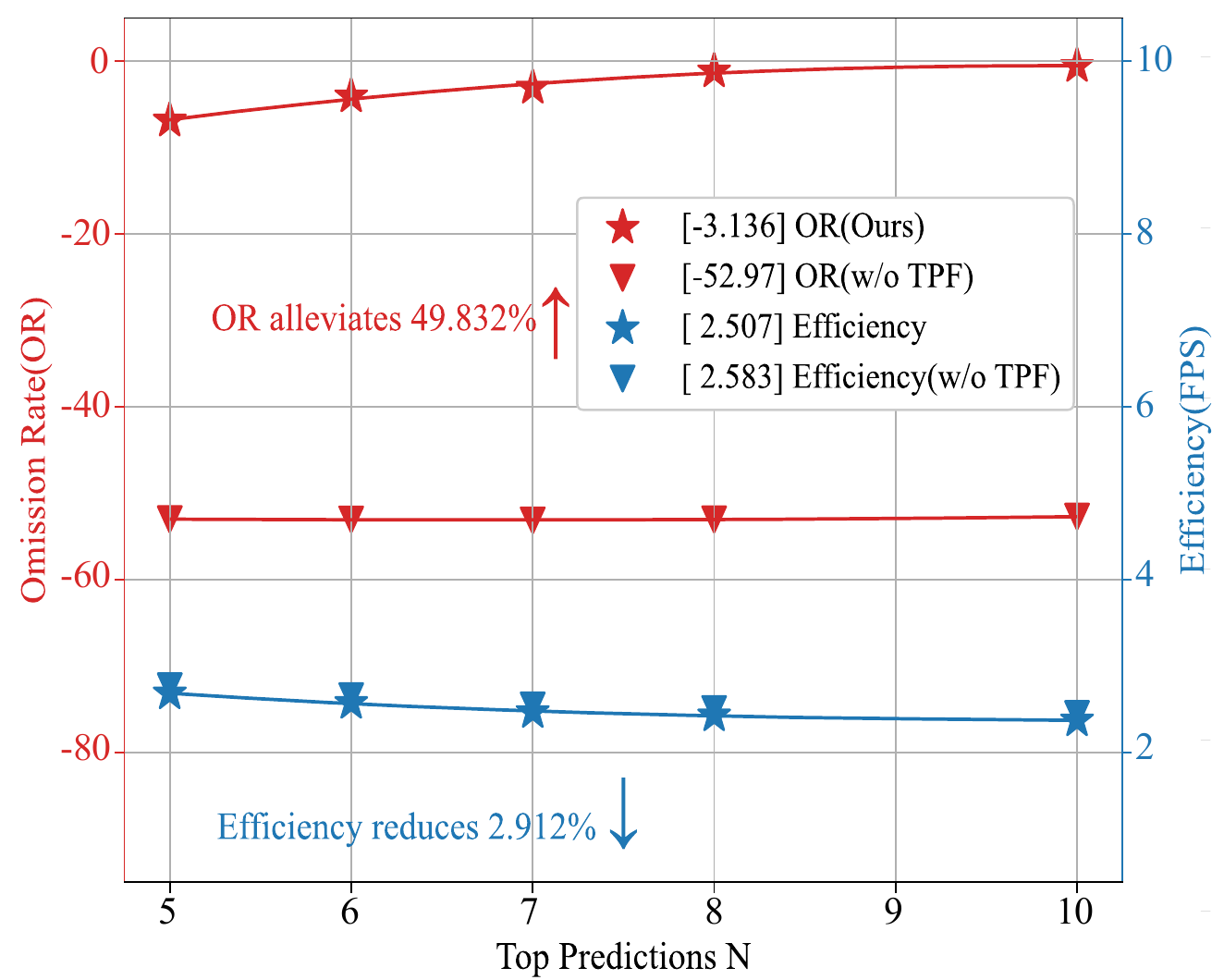}
    }
    \caption{Ablation of TPF module regarding OR and efficiency (K=3)}
    \label{OOR}
    
\end{figure}

\subsection{Efficient Inference} 
In this section, all the experiments run with a minor loop to evaluate the effectiveness of AirShot. We randomly sample the score for each category as the baseline, denoted as \textit{not applying TPF}. We found that the inference time is invariant to the shot settings, so we only report the inference time of different top prediction settings. We first show the result of applying AirShot with the Top 10 strategy in Table~\ref{eff}.

\begin{table}[ht]
\centering
\caption{The effect of AirShot applying Top N strategy(N=10)}
\resizebox{\linewidth}{!}{
\begin{tabular}{c ccccccc}
\toprule
Dataset & TPF & $OR_{1}$ & $ OR_{2} $ & $OR_{3}$& $OR_{5}$ & $OR_{\text{avg}}$ & T(s) \\
\midrule
Base & - & 0.00 & 0.00 & 0.00 & 0.00 & 0.00 & 0.733 \\
\midrule
COCO & \XSolidBrush & -55.9 & -54.5 & -48.7 & -51.7 & -52.7 & 0.388\\
COCO & \Checkmark & \textbf{-3.12} & \textbf{-1.71} & \textbf{-1.51} & \textbf{-1.50} & \textbf{-1.96} & 0.392\\
VOC & \XSolidBrush & -51.07 & -49.74 & -46.7 & -50.9 & -49.6 & 0.382\\
VOC & \Checkmark & \textbf{-4.07} & \textbf{-3.74} & \textbf{-1.95} & \textbf{-2.10} & \textbf{-2.96} & 0.386\\
\bottomrule
\end{tabular}}
\label{eff}
\end{table}

\begin{figure}[ht]
    \centering
    \resizebox{\linewidth}{!}{
    \includegraphics{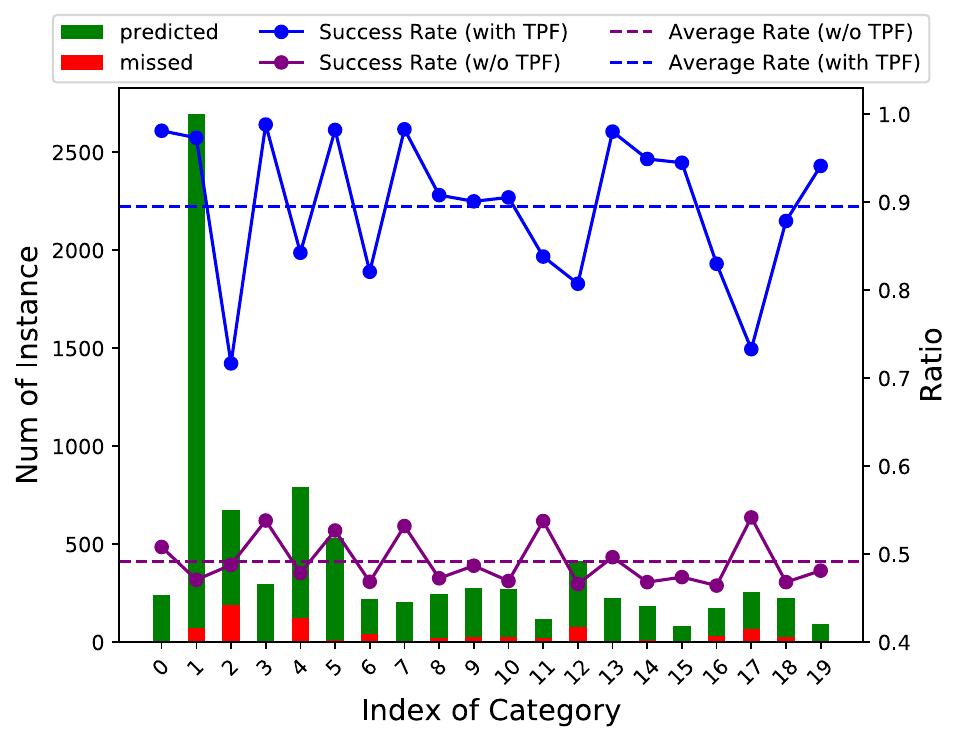}
    }
    \caption{Recall of TPF module. The \textbf{green} bar represents successful prediction by TPF, while the \textbf{red} part shows the missed instance. We also report the success rate in class-wise and averaged manner.}
    \label{recall}
\end{figure}

We also compare the efficiency without TPF to illustrate the inherent computational cost of TPF is negligible, as in Fig.~\ref{OOR}.
We use the ground truth category label in COCO to evaluate the recall rate of the TPF module and visualize as Fig.~\ref{recall}. The table shows that AirShot accurately infers the existence among most categories with an average recall rate of nearly 90\%. We highlight that any pre-selection methods cannot achieve 100\% percent. Since many false predictions are led by low-quality correlation maps, omission caused by this reason will not cause any performance degradation. 
\subsection{Inter-Model Comparison}
\begin{figure}[h]
    \centering
    \resizebox{\linewidth}{!}{
    \includegraphics[width=\textwidth]{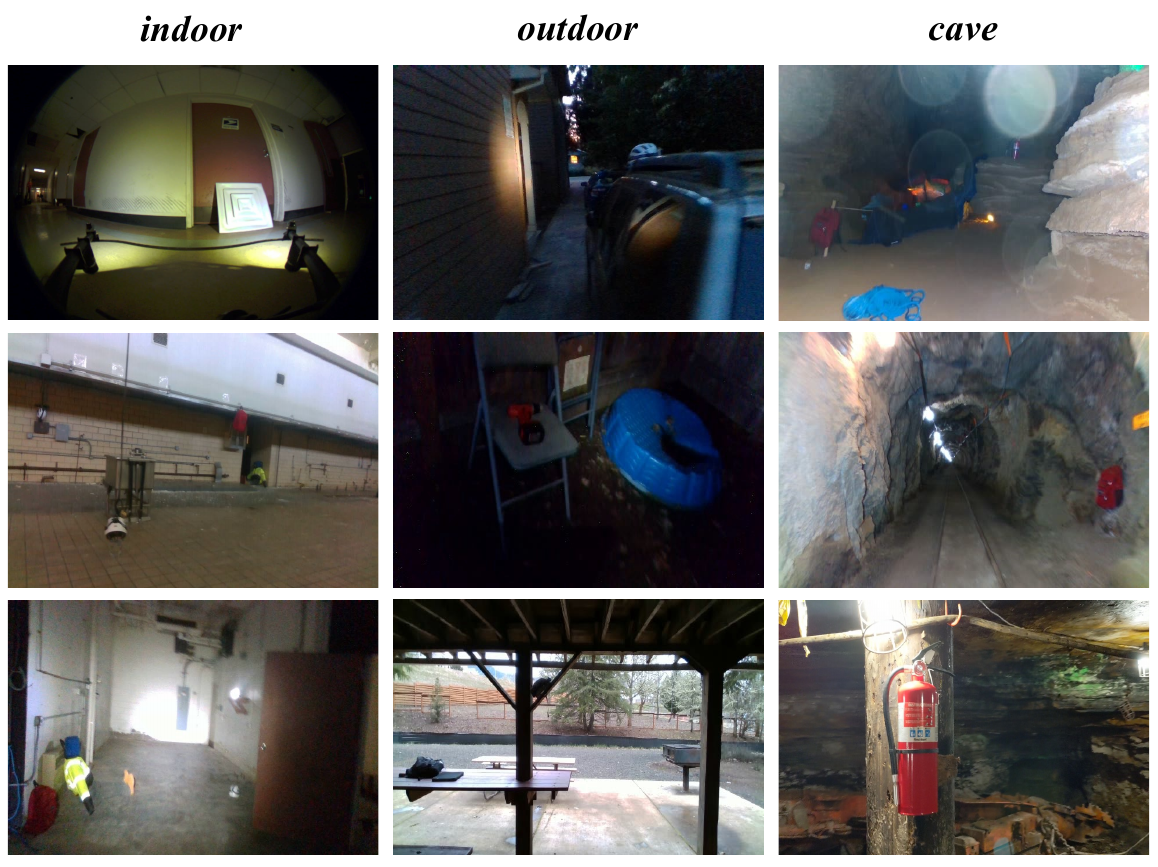}
    }
    \caption{Preview of our proposed dataset from the DARPA SubT Challenge, showcasing its reality, complexity, and diversity across various scenarios including indoor, outdoor, and cave environments.}
    \label{inter-mmodel}
\end{figure}

We choose all models that can work without fine-tuning stage for comparison in Fig.~\ref{inter-mmodel}. We also extend different top prediction settings to show our superior performance. Here we regard $N$ as a hyperparameter where $N$ ranges from 5 to 10. Then we realized that the number of novel classes on one query image varies, thus we empirically set a threshold enabling the adaptive selecting process. Due to unsupported filter characteristics, A-RPN and AirDet run a full inference. AirShot runs a minor inference loop powered by TPF. The result shows that our model significantly beats others in terms of precision and efficiency. One noticeable thing is adaptive method lies beyond the formed line by the Top-$N$ strategy. Although the adaptive method cannot achieve both the best performance and efficiency, it still shows a better balance. In our experiment, the threshold value is set empirically.

\begin{figure}[t]
    \centering
    % Adjust both the width and height while preserving the aspect ratio
    \resizebox{\linewidth}{!}{
    \includegraphics{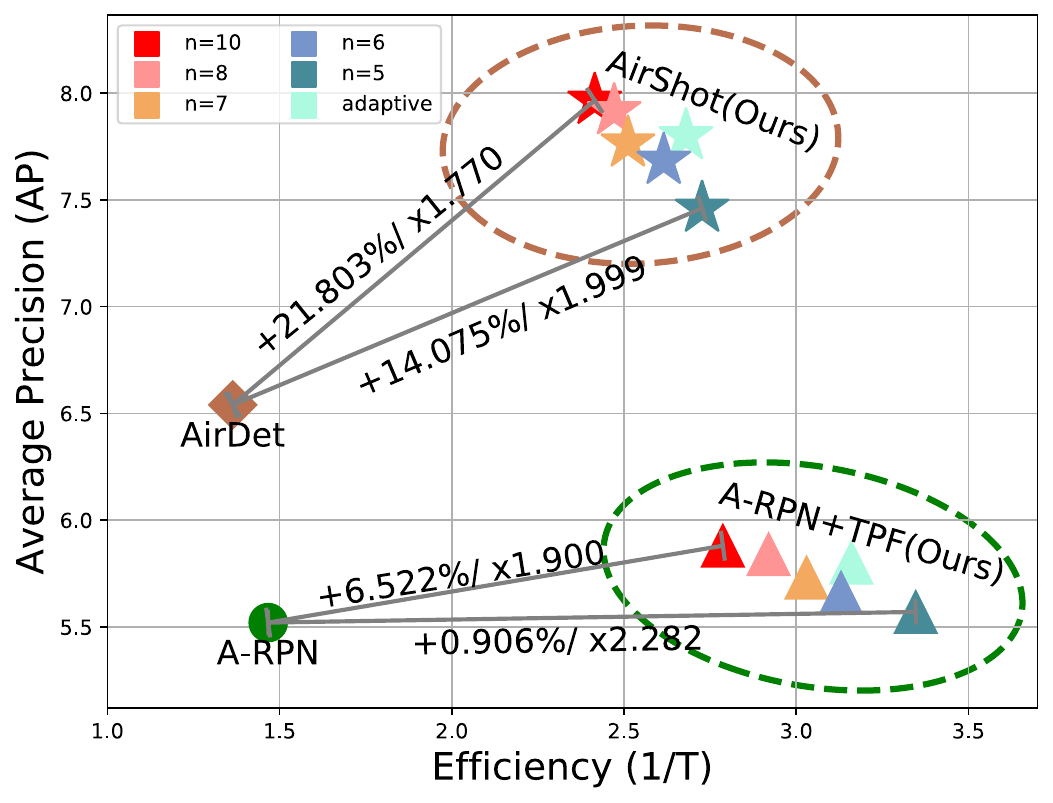}
    }
    \caption{Inter-model comparison among fine-tuning free models.}
    \label{inter-model}
\end{figure}

\subsection{Dataset Contribution and Real-world Test}
We collect and open-source a real dataset from DARPA Subterranean (SubT) challenge~\cite{subtchallenge} The environments poses extra difficulties, e.g., a lack of lighting, dripping water, and cluttered or irregularly shaped environments, etc. 

To test AirShot in real world, we adopted it on SubT where the robot is equipped with an NVIDIA Jetson AGX Xavier. Dataset preview and qualitative result are presented in supplementary materials. The effectiveness and efficiency of AirShot in the real-world tests demonstrated its promising prospect and feasibility for autonomous exploration.

\subsection{Ablation Study and Deep Visualizations}

\subsubsection{Visualization}
\begin{figure}[ht]
    \centering
    \resizebox{\linewidth}{!}{
    \includegraphics{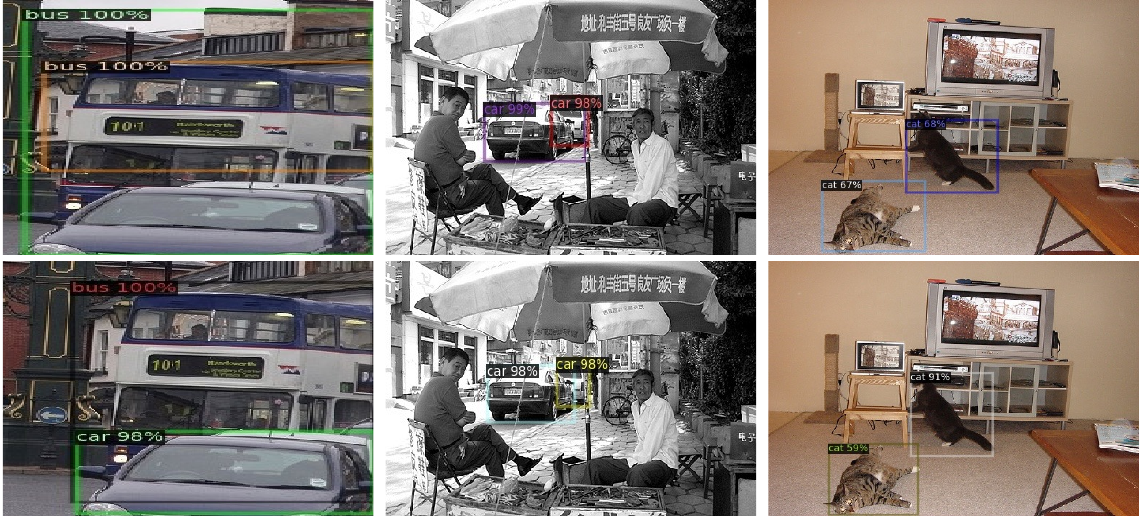}
    }

    \caption{Deep visualization of qualitatively comparison between previous best model \textbf{Top:} AirDet (baseline) and \textbf{Bottom}: Airshot (ours). AirShot can focus more precisely on the most representative part of the object, resulting in more accurate box regression.}
    \label{detection-result}
\end{figure}

We validate the effectiveness of AirShot by showing
qualitative visualization with detection results~\ref{detection-result}.

% In this section, we adopt a default test setting of 3-shot on COCOval 2014 dataset.
% \subsubsection{Ablation Study}
\subsubsection{Level of Correlation Maps}
The level of correlation maps is crucial during training and inference as they contain different levels of mutual information. Our choices are res2, res3, res4, and the fused correlation maps as in Table~\ref{different  correlation}:

\begin{table}[ht]
\centering
\caption{Effect of different level correlation maps tested in coco}
\begin{tabular}{c|c|c|c}
\toprule
method & training & inference & OR \\
\midrule
& \textbf{fused} & \textbf{res4} & \textbf{-1.51} \\
without TPF(K=20) & fused & fused & -2.54 \\
& res4 & fused & -6.29 \\
\midrule

 & res4 & res4 & -4.83 \\
with TPF(K=10) & fused & res3 & -26.0 \\
 & fused & res2 & -58.7 \\
\bottomrule
\end{tabular}
\label{different  correlation}
\end{table}

We surprisingly found that the fused correlation map is the sub-optimal during inference stage, the finest feature actually could help the TPF to be more discriminative in determining the logits for existence. Another strength of this property is that it skips the time-consuming feature fusion stage for those classes that are unlikely to appear in the query images, which increases the inference efficiency further.

\subsubsection{Representations}

To illustrate the necessity of our proposed global and local representations, we conduct the ablation study for those two branches as shown in Table~\ref{Representations_Global}.
The results illustrate that both global and local representations are crucial to a successful mapping building to fully extract information directly from correlation maps.

\begin{table}[ht]
\centering
\caption{Ablation Study of Different Representations}
\begin{tabular}{c|c|c|c|c|c}
\toprule
Global & Local & OR & AP & AP50 &AP75   \\
\midrule
 \XSolidBrush & \Checkmark & -9.75 & 7.22  & 13.50 & 7.11\\
 \Checkmark & \XSolidBrush & -6.14 & 7.51 & 14.04 & 7.36 \\
 \Checkmark & \Checkmark & \textbf{-1.51} & \textbf{8.01} & \textbf{14.96} & \textbf{7.87}  \\
\bottomrule
\end{tabular}
\label{Representations_Global}
\end{table}

\section{Conclusions \& Limitations}

We present a novel few-shot object detection system, AirShot for autonomous exploration of mobile robots. Specifically, AirShot fully exploits the correlation map for a more robust and faster FSOD system. It offers a dual functionality that substantially improves the efficiency and effectiveness of most off-the-shelf models. In the experiments, we show that it achieves state-of-the-art performance and efficiency among current FSOD models that can work without fine-tuning. Additionally, it is proven to be generalized efficiently in various meta-learning-based methods. We expect this method to play an important role in various robotic applications.

However, it also has several limitations: (1) We adopt a relatively simple structure to avoid extra computational cost thus there is still a gap between strictly precise rearranging score order; (2) We use the same model structure in the training stage, while other complicated models can perform better; and (3) When the number of novel classes is close to validation set, the enhancement would be limited.

\addtolength{\textheight}{-0cm}   % This command serves to balance the column lengths

%%%%%%%%%%%%%%%%%%%%%%%%%%%%%%%%%%%%%%%%%%%%%%%%%%%%%%%%%%%%%%%%%%%%%%%%%%%%%%%%

%%%%%%%%%%%%%%%%%%%%%%%%%%%%%%%%%%%%%%%%%%%%%%%%%%%%%%%%%%%%%%%%%%%%%%%%%%%%%%%%

%%%%%%%%%%%%%%%%%%%%%%%%%%%%%%%%%%%%%%%%%%%%%%%%%%%%%%%%%%%%%%%%%%%%%%%%%%%%%%%%

%%%%%%%%%%%%%%%%%%%%%%%%%%%%%%%%%%%%%%%%%%%%%%%%%%%%%%%%%%%%%%%%%%%%%%%%%%%%%%%%

%%%%%%%%% REFERENCES
\bibliographystyle{IEEEtran}
\bibliography{./egbib.bib}

\begin{thebibliography}{10}
\providecommand{\url}[1]{#1}
\csname url@samestyle\endcsname
\providecommand{\newblock}{\relax}
\providecommand{\bibinfo}[2]{#2}
\providecommand{\BIBentrySTDinterwordspacing}{\spaceskip=0pt\relax}
\providecommand{\BIBentryALTinterwordstretchfactor}{4}
\providecommand{\BIBentryALTinterwordspacing}{\spaceskip=\fontdimen2\font plus
\BIBentryALTinterwordstretchfactor\fontdimen3\font minus \fontdimen4\font\relax}
\providecommand{\BIBforeignlanguage}[2]{{%
\expandafter\ifx\csname l@#1\endcsname\relax
\typeout{** WARNING: IEEEtran.bst: No hyphenation pattern has been}%
\typeout{** loaded for the language `#1'. Using the pattern for}%
\typeout{** the default language instead.}%
\else
\language=\csname l@#1\endcsname
\fi
#2}}
\providecommand{\BIBdecl}{\relax}
\BIBdecl

\bibitem{fan2020few}
Q.~Fan, W.~Zhuo, C.-K. Tang, and Y.-W. Tai, ``{Few-Shot Object Detection with Attention-RPN and Multi-Relation Detector},'' in \emph{Proceedings of the IEEE/CVF Conference on Computer Vision and Pattern Recognition (CVPR)}, 2020, pp. 4013--4022.

\bibitem{Hu2021CVPR}
H.~Hu, S.~Bai, A.~Li, J.~Cui, and L.~Wang, ``{Dense Relation Distillation with Context-aware Aggregation for Few-Shot Object Detection},'' in \emph{Proceedings of the IEEE/CVF Conference on Computer Vision and Pattern Recognition (CVPR)}, 2021, pp. 10\,185--10\,194.

\bibitem{kang2019few}
B.~Kang, Z.~Liu, X.~Wang, F.~Yu, J.~Feng, and T.~Darrell, ``{Few-Shot Object Detection via Feature Reweighting},'' in \emph{Proceedings of the IEEE/CVF International Conference on Computer Vision (ICCV)}, 2019, pp. 8420--8429.

\bibitem{wang2019meta}
Y.-X. Wang, D.~Ramanan, and M.~Hebert, ``{Meta-Learning to Detect Rare Objects},'' in \emph{Proceedings of the IEEE/CVF International Conference on Computer Vision (ICCV)}, 2019, pp. 9925--9934.

\bibitem{wang2020frustratingly}
X.~Wang, T.~E. Huang, T.~Darrell, J.~E. Gonzalez, and F.~Yu, ``Frustratingly simple few-shot object detection,'' \emph{arXiv preprint arXiv:2003.06957}, 2020.

\bibitem{chen_tro}
C.~Wang, Y.~Qiu, W.~Wang, Y.~Hu, S.~Kim, and S.~Scherer, ``{Unsupervised Online Learning for Robotic Interestingness with Visual Memory},'' \emph{IEEE Transactions on Robotics}, pp. 1--15, 2021.

\bibitem{wang2020visual}
C.~Wang, W.~Wang, Y.~Qiu, Y.~Hu, and S.~Scherer, ``{Visual Memorability for Robotic Interestingness via Unsupervised Online Learning},'' in \emph{Proceedings of the European Conference on Computer Vision (ECCV)}, 2020, pp. 52--68.

\bibitem{li2022airdet}
B.~Li, C.~Wang, P.~Reddy, S.~Kim, and S.~Scherer, ``Airdet: Few-shot detection without fine-tuning for autonomous exploration,'' in \emph{Computer Vision--ECCV 2022: 17th European Conference, Tel Aviv, Israel, October 23--27, 2022, Proceedings, Part XXXIX}.\hskip 1em plus 0.5em minus 0.4em\relax Springer, 2022, pp. 427--444.

\bibitem{kim2022robotic}
S.~Kim, C.~Wang, B.~Li, and S.~Scherer, ``{Robotic Interestingness via Human-Informed Few-Shot Object Detection},'' in \emph{Proceedings of the IEEE/RSJ International Conference on Intelligent Robots and Systems (IROS)}, 2022, pp. 1756--1763.

\bibitem{li2021few}
Y.~Li, H.~Zhu, Y.~Cheng, W.~Wang, C.~S. Teo, C.~Xiang, P.~Vadakkepat, and T.~H. Lee, ``Few-shot object detection via classification refinement and distractor retreatment,'' in \emph{Proceedings of the IEEE/CVF Conference on Computer Vision and Pattern Recognition}, 2021, pp. 15\,395--15\,403.

\bibitem{yang2022efficient}
Z.~Yang, C.~Zhang, R.~Li, Y.~Xu, and G.~Lin, ``Efficient few-shot object detection via knowledge inheritance,'' \emph{IEEE Transactions on Image Processing}, vol.~32, pp. 321--334, 2022.

\bibitem{fan2021generalized}
Z.~Fan, Y.~Ma, Z.~Li, and J.~Sun, ``Generalized few-shot object detection without forgetting,'' in \emph{Proceedings of the IEEE/CVF Conference on Computer Vision and Pattern Recognition}, 2021, pp. 4527--4536.

\bibitem{qiao2021defrcn}
L.~Qiao, Y.~Zhao, Z.~Li, X.~Qiu, J.~Wu, and C.~Zhang, ``Defrcn: Decoupled faster r-cnn for few-shot object detection,'' in \emph{Proceedings of the IEEE/CVF International Conference on Computer Vision}, 2021, pp. 8681--8690.

\bibitem{sun2021fsce}
B.~Sun, B.~Li, S.~Cai, Y.~Yuan, and C.~Zhang, ``{FSCE: Few-Shot Object Detection via Contrastive Proposal Encoding},'' in \emph{Proceedings of the IEEE/CVF Conference on Computer Vision and Pattern Recognition (CVPR)}, 2021, pp. 7352--7362.

\bibitem{wu2020multi}
J.~Wu, S.~Liu, D.~Huang, and Y.~Wang, ``{Multi-Scale Positive Sample Refinement for Few-Shot Object Detection},'' in \emph{Proceedings of the European Conference on Computer Vision (ECCV)}, 2020, pp. 456--472.

\bibitem{zhang2021accurate}
L.~Zhang, S.~Zhou, J.~Guan, and J.~Zhang, ``{Accurate Few-Shot Object Detection With Support-Query Mutual Guidance and Hybrid Loss},'' in \emph{Proceedings of the IEEE/CVF Conference on Computer Vision and Pattern Recognition (CVPR)}, 2021, pp. 14\,424--14\,432.

\bibitem{zhang2020cooperating}
W.~Zhang, Y.-X. Wang, and D.~A. Forsyth, ``Cooperating rpn's improve few-shot object detection,'' \emph{arXiv preprint arXiv:2011.10142}, 2020.

\bibitem{hadsell2006dimensionality}
R.~Hadsell, S.~Chopra, and Y.~LeCun, ``Dimensionality reduction by learning an invariant mapping,'' in \emph{2006 IEEE computer society conference on computer vision and pattern recognition (CVPR'06)}, vol.~2.\hskip 1em plus 0.5em minus 0.4em\relax IEEE, 2006, pp. 1735--1742.

\bibitem{girshick2015fast}
R.~Girshick, ``Fast r-cnn,'' in \emph{Proceedings of the IEEE international conference on computer vision}, 2015, pp. 1440--1448.

\bibitem{girshick2014rich}
R.~Girshick, J.~Donahue, T.~Darrell, and J.~Malik, ``Rich feature hierarchies for accurate object detection and semantic segmentation,'' in \emph{Proceedings of the IEEE conference on computer vision and pattern recognition}, 2014, pp. 580--587.

\bibitem{liu2016ssd}
W.~Liu, D.~Anguelov, D.~Erhan, C.~Szegedy, S.~Reed, C.-Y. Fu, and A.~C. Berg, ``{SSD: Single Shot Multibox Detector},'' in \emph{Proceedings of the European Conference on Computer Vision (ECCV)}, 2016, pp. 21--37.

\bibitem{redmon2017yolo9000}
J.~Redmon and A.~Farhadi, ``Yolo9000: better, faster, stronger,'' in \emph{Proceedings of the IEEE conference on computer vision and pattern recognition}, 2017, pp. 7263--7271.

\bibitem{ren2015faster}
S.~Ren, K.~He, R.~Girshick, and J.~Sun, ``Faster r-cnn: Towards real-time object detection with region proposal networks,'' \emph{Advances in neural information processing systems}, vol.~28, 2015.

\bibitem{redmon2016you}
J.~Redmon, S.~Divvala, R.~Girshick, and A.~Farhadi, ``You only look once: Unified, real-time object detection,'' in \emph{Proceedings of the IEEE conference on computer vision and pattern recognition}, 2016, pp. 779--788.

\bibitem{redmon2018yolov3}
J.~Redmon and A.~Farhadi, ``Yolov3: An incremental improvement,'' \emph{arXiv preprint arXiv:1804.02767}, 2018.

\bibitem{lin2017fpn}
T.-Y. Lin, P.~Doll{\'a}r, R.~Girshick, K.~He, B.~Hariharan, and S.~Belongie, ``{Feature Pyramid Networks for Object Detection},'' in \emph{Proceedings of the IEEE/CVF Conference on Computer Vision and Pattern Recognition (CVPR)}, 2017, pp. 2117--2125.

\bibitem{yan2019meta}
X.~Yan, Z.~Chen, A.~Xu, X.~Wang, X.~Liang, and L.~Lin, ``{Meta R-CNN: Towards General Solver for Instance-Level Low-Shot Learning},'' in \emph{Proceedings of the IEEE/CVF International Conference on Computer Vision (ICCV)}, 2019, pp. 9577--9586.

\bibitem{han2021query}
G.~Han, Y.~He, S.~Huang, J.~Ma, and S.-F. Chang, ``Query adaptive few-shot object detection with heterogeneous graph convolutional networks,'' in \emph{Proceedings of the IEEE/CVF International Conference on Computer Vision}, 2021, pp. 3263--3272.

\bibitem{zhu2021semantic}
C.~Zhu, F.~Chen, U.~Ahmed, Z.~Shen, and M.~Savvides, ``{Semantic Relation Reasoning for Shot-Stable Few-Shot Object Detection},'' in \emph{Proceedings of the IEEE/CVF Conference on Computer Vision and Pattern Recognition (CVPR)}, 2021, pp. 8782--8791.

\bibitem{wu2021universal}
A.~Wu, Y.~Han, L.~Zhu, and Y.~Yang, ``Universal-prototype enhancing for few-shot object detection,'' in \emph{Proceedings of the IEEE/CVF International Conference on Computer Vision}, 2021, pp. 9567--9576.

\bibitem{zhang2022meta}
G.~Zhang, Z.~Luo, K.~Cui, S.~Lu, and E.~P. Xing, ``Meta-detr: Image-level few-shot detection with inter-class correlation exploitation,'' \emph{IEEE Transactions on Pattern Analysis and Machine Intelligence}, 2022.

\bibitem{han2022few}
G.~Han, J.~Ma, S.~Huang, L.~Chen, and S.-F. Chang, ``Few-shot object detection with fully cross-transformer,'' in \emph{Proceedings of the IEEE/CVF conference on computer vision and pattern recognition}, 2022, pp. 5321--5330.

\bibitem{Kong2016HyperNet}
T.~Kong, A.~Yao, Y.~Chen, and F.~Sun, ``{HyperNet: Towards Accurate Region Proposal Generation and Joint Object Detection},'' in \emph{Proceedings of the IEEE/CVF Conference on Computer Vision and Pattern Recognition (CVPR)}, June 2016.

\bibitem{li2017fssd}
Z.~Li and F.~Zhou, ``{FSSD: Feature Fusion Single Shot Multibox DAetector},'' \emph{arXiv preprint arXiv:1712.00960}, 2017.

\bibitem{Shen2017dsod}
Z.~Shen, Z.~Liu, J.~Li, Y.-G. Jiang, Y.~Chen, and X.~Xue, ``{DSOD: Learning Deeply Supervised Object Detectors From Scratch},'' in \emph{Proceedings of the IEEE International Conference on Computer Vision (ICCV)}, Oct 2017.

\bibitem{he2016deep}
K.~He, X.~Zhang, S.~Ren, and J.~Sun, ``Deep residual learning for image recognition,'' in \emph{Proceedings of the IEEE conference on computer vision and pattern recognition}, 2016, pp. 770--778.

\bibitem{subtchallenge}
\url{https://subtchallenge.com}.

\end{thebibliography}

\end{document}